\documentclass{article}

\PassOptionsToPackage{numbers}{natbib}
\usepackage[preprint]{neurips_2023}

\usepackage{microtype}
\usepackage{graphicx}
\usepackage{subfigure}
\usepackage{booktabs} 
\usepackage{multirow}

\usepackage{amsmath}
\usepackage{amssymb}
\usepackage{mathtools}
\usepackage{amsthm}

\usepackage[utf8]{inputenc} 
\usepackage[T1]{fontenc}    
\usepackage{hyperref}       
\usepackage[capitalise]{cleveref}
\usepackage{url}            
\usepackage{booktabs}       
\usepackage{amsfonts}       
\usepackage{nicefrac}       
\usepackage{microtype}      
\usepackage{xcolor}         

\DeclareMathOperator{\FAT}{FAT}
\DeclareMathOperator*{\argmax}{arg\,max}

\title{Fully Embedded Time-Series Generative Adversarial Networks}

 \author{%
   Joe Beck \\
   \texttt{jbeck9@vols.utk.edu} \\
   University of Tennessee \\ 
   Department of Mechanical, Aerospace, \& Biomedical Engineering \\ 
   Knoxville, TN, USA\\
   ORCID: 0000-0003-4257-1138
   \And
   Subhadeep Chakraborty \\
   \texttt{schakrab@utk.edu} (Corresponding author)\\
   University of Tennessee \\ 
   Department of Mechanical, Aerospace, \& Biomedical Engineering \\ 
   Knoxville, TN, USA\\
   ORCID: 0000-0001-5035-9925}

\begin{document}

\newif\ifrevoneflag 
\revoneflagfalse
\newcommand{\revone}[1]{\ifrevoneflag\color{blue}#1\color{black}\else#1\fi}

\newcommand\xot{x_{1:T}}
\newcommand\Xot{X_{1:T}}
\newcommand\E{\mathbb{E}}

\maketitle

 \begin{abstract}
   Generative Adversarial Networks (GANs) should produce synthetic data that fits the underlying distribution of the data being modeled. For real valued time-series data, this implies the need to simultaneously capture the \textit{static} distribution of the data, but also the full \textit{temporal} distribution of the data for any potential time horizon. This temporal element produces a more complex problem that can potentially leave current solutions under-constrained, unstable during training, or prone to varying degrees of mode collapse. In \textit{FETSGAN}, entire sequences are translated directly to the generator’s sampling space using a \textit{seq2seq} style adversarial autoencoder (AAE), where adversarial training is used to match the training distribution in both the feature space and the lower dimensional sampling space. This additional constraint provides a loose assurance that the temporal distribution of the synthetic samples will not collapse. In addition, the First Above Threshold (FAT) operator is introduced to supplement the reconstruction of encoded sequences, which improves training stability and the overall quality of the synthetic data being generated. These novel contributions demonstrate a significant improvement to the current state of the art for adversarial learners in qualitative measures of temporal similarity and quantitative predictive ability of data generated through \textit{FETSGAN}.

   \textbf{Keywords}:  Generative Adversarial Networks (GANs), adversarial autoencoder, synthetic time series data 
 \end{abstract}

 \section*{Statements and Declarations}
 \subsection*{Competing Interests}
 The authors declare no known competing interests in the content of this manuscript. 

 \subsection*{Funding}
 Funding for this work was partially provided by the Collaborative Sciences Center for Road Safety (CSCRS), as well as the University of Tennessee, Knoxville.

 \newpage

\section{Introduction}
Generative modeling is the field of research concerned with producing new and unique data that is similar to the data that was used to produce the model. More specifically, we can say that this similarity is defined in terms of the ability to model the underlying distribution represented by the training data. In the case of sequential vector $x_{1:T}$ with length $T$, the data distribution is characterized by the temporal distribution $p(x_1, ... ,x_T)$. Even with relatively simple datasets where the vector $x_t$ is low dimensional, compounding dependencies in the temporal distribution increase with $T$ until it becomes difficult to measure or even visualize the similarity or differences between the temporal distributions of the training data and the generated data. Generative Adversarial Networks (GANs) have demonstrated exceptional ability in modeling complex distributions such as these. However, GANs are notoriously difficult to train, with instability often preventing convergence and the final generative models featuring some degree of mode collapse, where only a portion of the full target distribution is represented in the synthetic samples.

\begin{figure*}[h]
\includegraphics[width=\textwidth]{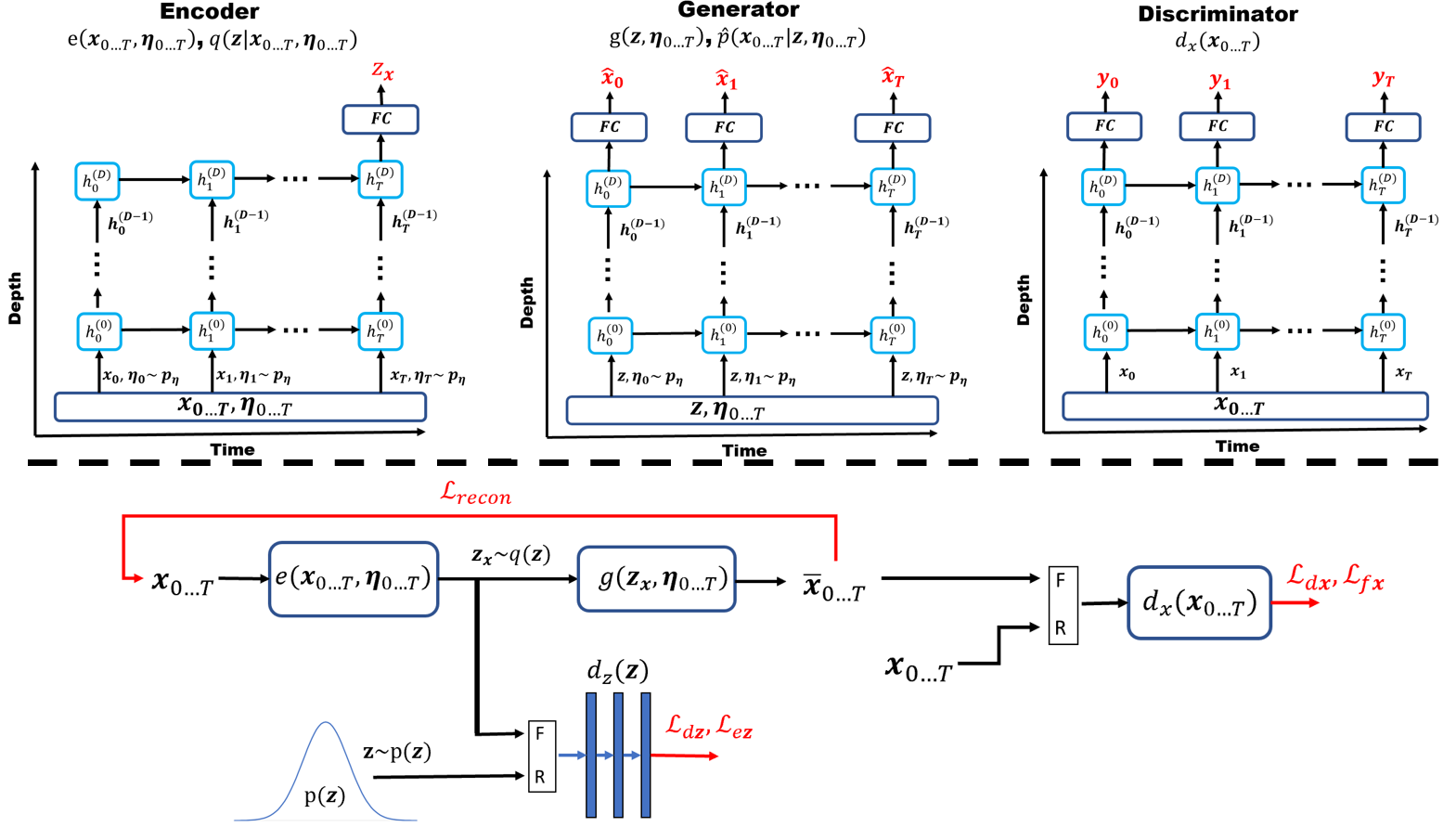}
\caption{The overall training scheme is shown. Above the dashed line, the RNN style architectures are detailed, showing the model outputs (red) of each network as a function of the inputs. $h^{(d)}_t$ indicates a hidden state of the network, where $d$ represents the weights associated with a specific depth and $t$ represents that state any specific time. FC indicates a fully connected output layer, and these weights are shared for every output across time, i.e. in the case of the generator $\hat{{x}}_{1:T}$. Below the dashed line, the training flow is visualized. The mechanisms for producing \textit{FETSGAN}'s five objective functions are shown in red. 
}
\label{overview}
\centering
\end{figure*}

Like much of the work surrounding GANs, the novel process presented here provides additional constraints to the adversarial learning process that regularizes learning, resulting in greater stability during training, higher quality data, and less susceptibility to mode collapse, specifically in the temporal distributions. The architecture presented is a modification of RCGAN\cite{estebanRealvaluedMedicalTime2017} or C-RNN-GAN\cite{mogrenCRNNGANContinuousRecurrent2016} that features a \textit{seq2seq} style AAE as encoder and decoder for data generation \cite{makhzaniAdversarialAutoencoders2016}, \cite{sutskeverSequenceSequenceLearning2014a}. The complete model is visualized in \cref{overview}. There are three primary benefits to using an adversarial autoencoder. First, there is an additional constraint that matches the posterior distribution of the encodings to the prior distribution of the samples at inference time, further combating mode collapse beyond feature-level adversarial training. Second, the \textit{seq2seq} encoder is forced to summarize the entire sequence at once, allowing it to capture relevant time dependencies of arbitrary length. This can be compared to the approach in the TimeGAN, where the regularizing effect of the teacher-forced supervised losses of the encodings are only applied one time-step into the future \cite{yoonTimeseriesGenerativeAdversarial2019}. Finally, the posterior encodings of adversarial autoencoders are natively interpretable. This allows fine control over the style of the synthetic data being generated, even in the completely unsupervised learning setting of this work.

The decoder in our framework can potentially suffer from the same supervised training problems as any other autoregressive model where compounding errors over time across shared weights can cause slow or unstable training, especially during the reconstruction of long encoded sequences. Typically this is resolved with some variation of teacher-forcing \cite{bengioScheduledSamplingSequence2015}, \cite{lambProfessorForcingNew2016}, where the network is provided the ground truth from the previous timestep, and learns to predict only one timestep into the future. This method often leaves the network with some degree of exposure bias, where the compounding error of the model’s own predictions are neglected. Additionally in the context of real-valued time-series generation, statistical properties in the data may produce harsh local minimums for regression based optimization. In this work, we do not use an autoregressive decoder at all. Instead, our solution to reconstruction loss is coined First Above Threshold (FAT) loss. In this scheme, stochastic gradient descent is only applied to the model parameters at one time instance per generated sequence. This allows the network to learn progressively longer sequences during training, and can be applied to any element-wise loss function in a supervised or unsupervised manner. 
The work described here is an extension of RCGAN and a reformulation of TimeGAN that utilizes a supervised loss in the feature space as well as an unsupervised loss on the encodings of the data which produces improved matching of the temporal distribution of the training data. In addition, FAT loss can improve the quality of the data being generated, as well as producing a standardizing effect on the training dynamics. For our experiments, stocks data, energy usage data, metro traffic data, and a synthetic sine wave dataset are used. The sequences produced are qualitatively analysed to show significant improvement in preventing mode collapse along the temporal distribution. In addition, we demonstrate the ability to selectively sample our model at inference time to produce realistic data of a specific style. Finally, we measure the performance of our model against the primary adversarial learners in terms of predictive and discriminative scores. \textit{FETSGAN} shows significant improvement to these methods in all stated dimensions of analysis.

\section{Related Work}
As a generative model, \textit{FETSGAN} builds on the adversarial training scheme  of RCGAN \cite{estebanRealvaluedMedicalTime2017} with mixed supervised and unsupervised learning elements, along with a Recurrent Neural Network (RNN) style architecture for the Encoder, Decoder, and Discriminator, which are common for many sequence modeling tasks \cite{chungEmpiricalEvaluationGated2014}. The work most closely matching this description is TimeGAN \cite{yoonTimeseriesGenerativeAdversarial2019}, where the key distinction is our use of an AAE for full time sequences, as opposed to the element-wise encodings in TimeGAN. There is also the application specific R-GAN which combined some time-series heuristic information like Fourier representations with a WGAN approach to produce synthetic energy consumption data \cite{arjovskyWassersteinGAN2017}, \cite{fekriGeneratingEnergyData2020}. COT-GAN is a recent approach to this problem that defines an adversarial loss through an expansion of the Sinkhorn Divergence into the time domain \cite{xuCOTGANGeneratingSequential2020}. Due to the inherent instability of adversarial training, \cite{jarrettTimeseriesGenerationContrastive2021} regularizes training with a contrastive model and a training scheme grounded in imitation learning. \revone{Finally, \cite{colettaConstrainedTimeSeriesGeneration2023} incorporates two separate approaches using diffusion models and more traditional constrained optimization techniques to produce time-series data.} The experimental comparison here is limited to methods that use the most straightforward approach of applying adversarial learning to the feature space, or latent encodings of the feature space. Namely, these models are TimeGAN and RCGAN. This is partly due to the lack of working implementations of the alternative approaches, and also to the similarity in approach with these methods, as well as the fact that these simpler adversarial approaches still seem to remain the preeminent method of time series data generation where such models are applied, \revone{with applications in fields such as medicine \cite{dashMedicalTimeSeriesData2020, liCausalRecurrentVariational2023}, energy management \cite{chattorajImprovingStabilityAdversarial2021, fochesatoUseConditionalTimeGAN2022}, and sensor simulation \cite{adibSyntheticECGSignal2023}. Additionally, the use of transformers \cite{vaswaniAttentionAllYou2017} has become very popular in the field of sequence generation, particularly with respect to large language models (LLMs) such as the popular Generative Pre-trained Transformer (GPT) architecture \cite{rayChatGPTComprehensiveReview2023}. While we did experiment with transformer-based approaches, we found that the method described here outperformed these methods. We postulate the difference in performance was due to transformer-based methods being auto-regressive in nature, exposing the model to inference bias that was not present in our non-auto-regressive approach.}

Beyond the direct comparisons with other models that produce realistic time-series data, \textit{FETSGAN} also incorporates an interpretive latent space, allowing the selective sampling of the posterior distribution at inference time to reach some desired effect, specified by the user. This bears a direct connection to the field of representation learning, where data is compressed in a meaningful way in order to accomplish some downstream task. This has been accomplished on real valued time series data, as data is embedded in RNN style architectures for the purpose of forecasting \cite{lyuImprovingClinicalPredictions2018}, supervised learning \cite{daiSemisupervisedSequenceLearning2015}, and data imputation \cite{bianchiLearningRepresentationsMultivariate2019}. Perhaps the most well-known interpretive generative models are variational autoencoders (VAEs) \cite{kingmaAutoEncodingVariationalBayes2022}. The alternative AAE used here has a close resemblance to this approach, as described in \cite{makhzaniAdversarialAutoencoders2016}, with the primary benefit in our use case being the ability to choose an arbitrary prior distribution instead of a standard Gaussian. The simplistic AAE used in this work has also been extended to utilize Wasserstein loss \cite{tolstikhinWassersteinAutoEncoders2019} in the image-generation space, demonstrating some improved stability that is typical of WGANs.

A theme of this work and all recent work on the creation of synthetic time-series is the regularization or complete abandonment of adversarial training as a means to produce more stability, and thus higher quality data. Obviously, this challenge is not unique to the time-series domain, and inspiration can be drawn from any generation process that utilizes adversarial learning, particularly image creation. Given the \textit{seq2seq} translation element of \textit{FETSGAN}, it stands to reason inspiration could be drawn from image-to-image translation techniques. We can see that the regularization effect of reconstruction loss or cycle loss is present in many adversarial approaches  \cite{liuUnsupervisedImagetoImageTranslation2017}, \cite{zhu2017unpaired}. WGANs \cite{arjovskyWassersteinGAN2017} and LSGANs \cite{maoLeastSquaresGenerative2017} focus primarily on the output layer of the discriminator and the loss function to produce a regularizing effect on reducing exploding or vanishing gradients. Spectral normalization \cite{miyatoSpectralNormalizationGenerative2018}, particularly in the weights of the discriminator, assures Lipchitz continuity in the gradients, further ensuring training stability. While not directly applicable to the RNN architecture itself, batch normalization in the discriminator has also demonstrated an ability to speed up adversarial training \cite{xiangEffectsBatchWeight2017}. Spectral normalization is utilized in the linear layers of the our discriminator’s training process, while reconstruction loss bares a resemblance to the cycle consistency loss that has a regularizing effect on training.

\section{Proposed Method}

In this section, the methodology of the proposed method is described in full detail. Additional motivation for the work is also provided.

\subsection{Problem Formulation}
Consider the random vector $X \in \mathcal{X}$, where individual instances are denoted by $x$. We operate in a discrete time setting where fixed time intervals exist between samples $X_t$, forming sequences of length $T$, such that $(X_1, ..., X_T) := \Xot \in \mathcal{X}^T$. We note that $T$ may be a constant value or may be a random variable itself, and that the proposed methodology does not differ in either case. The representative dataset of length $N$ is given by $\mathcal{D}=\{x_{n,1:T}\}_{n=1}^{N}$. For convenience, the subscript $n$ is omitted in further notation.

The data distribution $p(\Xot)$ describes the sequences, such that any possible conditional $p(X_t \mid X_{t-i})$ for $t \le T$ and $t-i \ge 0$ is absorbed into $p(\Xot)$. Thus, the overall learning objective is to produce the model distribution $\hat{p}(\Xot)$,
\begin{align}
\min_{\hat{p}} D(p(\Xot) \mid\mid \hat{p}(\Xot\ ))
\label{div1}
\end{align}\
where $D$ is some measure of divergence between the two distributions. We note here that RCGAN \cite{estebanRealvaluedMedicalTime2017} applies adversarial loss that minimizes this divergence directly. In the following works, higher quality data is generated by supplementing this loss function with a supervised loss that is more stable \cite{yoonTimeseriesGenerativeAdversarial2019}, or by avoiding adversarial loss altogether \cite{jarrettTimeseriesGenerationContrastive2021}. While the adversarial autoencoder in \cref{advauto} is intended to supplement and further stabilize training as well, we also highlight the additional ability of the $\FAT$ operator to stabilize the minimization of this objective on reconstructions in \cref{fat}.

\subsection{Adversarial Autoencoder}
\label{advauto}

In order to represent complex temporal relationships, we would like to leverage the ability to encode the entire time-series as a low dimensional vector in a \textit{seq2seq} style model. We introduce the random variable $Z \in \mathcal{Z}$ as an intermediate encoding for producing $\hat{p}(\Xot)$. We also introduce $\eta_{1:T}$ as a random noise vector from a known distribution $p_\eta$ sampled independent of time. Let $p_{z}(Z)$ be a known prior distribution, the encoder function $e:(\mathcal{X}^T,\mathcal{\eta}^T) \rightarrow \mathcal{Z}$ have the encoding distribution $q(Z\mid \Xot, \eta_{1:T})$, and the generator (or decoder) function $g: (\mathcal{Z},\mathcal{\eta}^T) \rightarrow \mathcal{X}^T$ have the decoding distribution $\hat{p}(\Xot\mid Z, \eta_{1:T})$.

Also, we can define an ideal mapping function that maps sequences to vectors $M: (\mathcal{X}^T,\eta^T) \rightarrow \mathcal{M}$, such that $M_X = M(\Xot, \eta_{1:T})$ and $p_{M}(M_X) = p(\Xot)p_{\eta}(\eta_{1:T})$. The aggregated posterior can now be defined,
\begin{align}
q(Z_X) = \int_{M_X} q(Z \mid \Xot, \eta_{1:T}) p_{M}(M_X) dM_X
\end{align}\
Similar to \cref{div1}, encoder training occurs by matching this aggregated posterior distribution $q(Z_X)$ to an arbitrary prior distribution $p_{Z}(Z)$.
\begin{align}
\min_{q} D(p_{z}(Z) \mid\mid q(Z_X))
\label{div2}
\end{align}\
The posterior distribution $q(Z_X)$ is equivalent to the universal approximator function in \cite{makhzaniAdversarialAutoencoders2016}, where the noise $\eta_t$ exists to provide stochasticity to the model in the event that not enough exists in the data $\mathcal{D}$ for $q(Z_X)$ to match $p_{z}(Z)$ deterministically. 

The generator is trained through reconstructing the data distribution $p(\Xot)$ while being conditioned on the encodings.
\begin{align}
\min_{\hat{p}} D(p(\Xot) \mid\mid \hat{p}(\Xot \mid Z_X, \eta_{1:T})))
\label{div3}
\end{align}\

If divergence is minimized in \cref{div2} and \cref{div3}, then the encoder is able to perfectly replicate the prior distribution, and the generator is perfectly able to reconstruct $\Xot$, resulting in a complete model of $p(\Xot)$ when the prior distribution $p_{z}(Z)$ is sampled and decoded at inference time.

\begin{figure*}[ht]
\begin{center}
\includegraphics[width=4in]{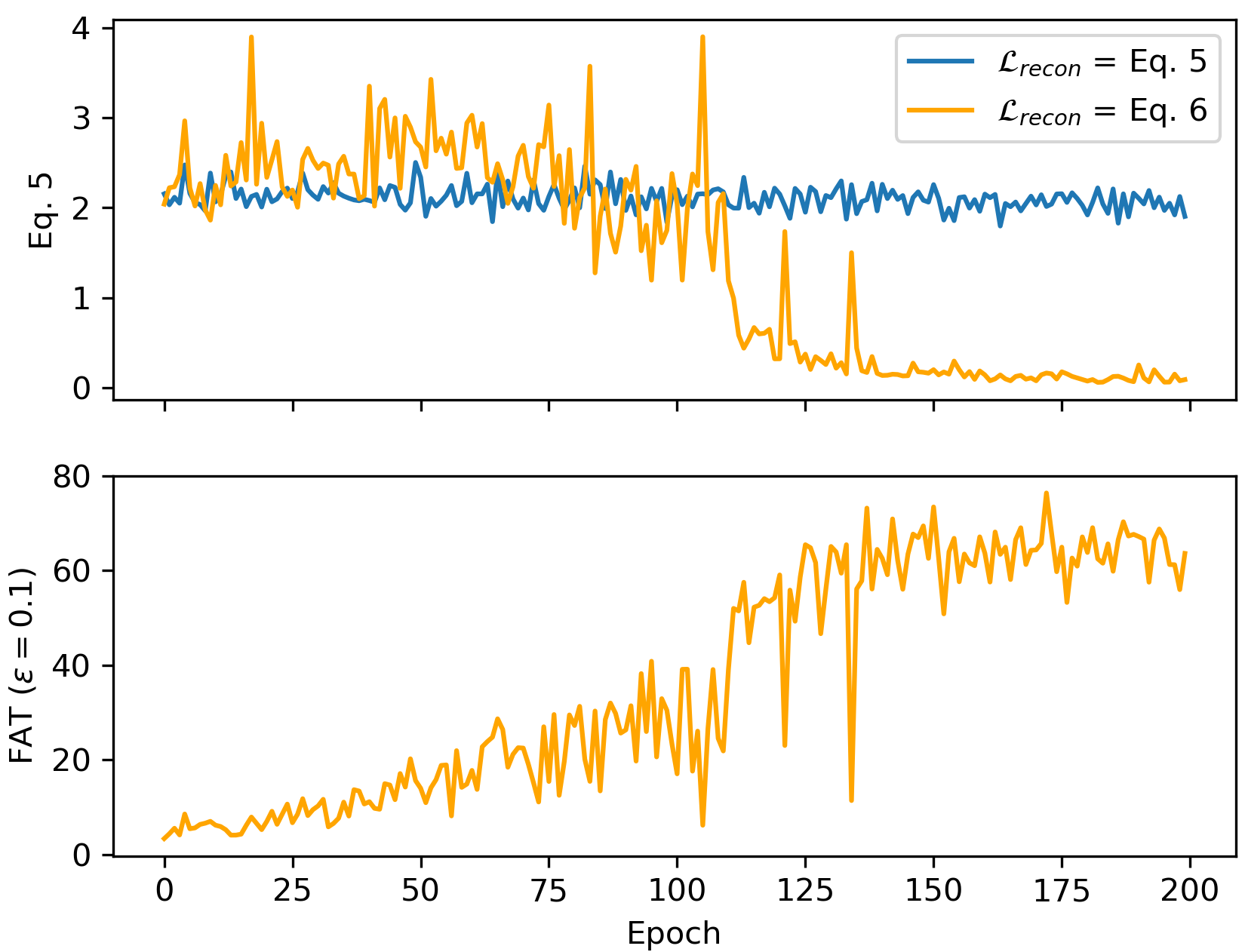}
\caption{For the encoding and reconstruction of $\bar{x}_{1:T} = g(e(\xot))$, the reconstruction objectives of \cref{sum_recon} and \cref{recon} are compared for the sines dataset. The average value for each epoch in $200$ epochs of training is plotted. Here, training occurs under the complete model objective of \cref{objective}. In the first row, we can see that the model is unable to apply adversarial learning and simultaneously learn the proper encodings to reconstruct $\xot$ under \cref{sum_recon}. This causes the optimization to immediately fall into a local minimum. With $\tau$ gradually increasing to progressively learn longer sequences in the second row, the $\FAT_{t}$ operation facilitates minimizing reconstruction loss better than applying \cref{sum_recon} directly.
}
\label{fatjust}
\end{center}
\end{figure*}

\subsection{First Above Threshold (FAT) Operator}
\label{fat}
In our formulation, the generator model minimizes two measures of divergence. First, there is the direct matching of the target distribution through adversarial training, characterized by \cref{div1}. Then, there is the secondary reconstruction loss from the intermediate encodings, characterized by \cref{div3}. Reconstructions of time series data are denoted $\bar{x}_{1:T} = g(e(\xot))$. The simplest loss function to enforce reconstruction of the original data might be,

\begin{align}
\mathcal{L}= \E_{\xot \sim p}(\sum_{t} \| x_{t}-\bar{x}_{t} \|^2)
\label{sum_recon}
\end{align}

where this objective is minimized in tandem by the generator and the encoder. This may prove to be challenging as our generator model is not an auto regressive model. It takes the intermediate encodings as input at every timestep, as shown in \cref{overview}. This reduces the possibility of a vanishing gradient to the encoder itself, and also prevents the generator from forgetting the initial encoding for long sequences. This does, however, make learning through reconstruction more difficult due to the inability to incorporate teacher forcing methods. In addition, time-series reconstruction with real-valued data faces the optimization problem of large local minimums, where the model may collapse into producing only mean values across any one time-series or the mean of the entire dataset. To alleviate these problems, we propose the solution of only applying reconstruction loss at \textit{one} time instance $t=\tau$ in the sequence, instead of the entire time series at once. Thus, the reconstruction loss,
\begin{align}
\mathcal{L}_{recon}= \E_{\xot \sim p}(\| x_{\tau}-\bar{x}_{\tau} \|^2)
\label{recon}
\end{align}
trains the generator in a supervised manner. The question remains which time instance $\tau$ to choose. We propose a simple solution. Prior to training, we define a real-valued threshold $\epsilon$, such that any error that has compounded to produce the reconstruction $\bar{x}_t$ is \textit{acceptable} so long as $\| x_{t}-\bar{x}_{t} \|^2 < \epsilon$. In this way, time-series reconstructions are progressively learned from short-term to long-term, instead of all at once. To this end, the First Above Threshold (FAT) operator is introduced. This operator takes as input a sequence $l_{1:T}$ and a threshold $\epsilon$, such that the minimum value for $t$ is returned where $l_{t} > \epsilon$. In the event that $l_t < \epsilon \, \forall \, t \in T$, $\FAT_{t}(l_{1:T}, \epsilon) = \argmax_{t}(l_{1:T})$. With this newly defined operator,
\begin{align}
\tau = \FAT_{t}(\| x_{t}-\bar{x}_{t} \|^2, \epsilon)
\label{taudef}
\end{align}
defines $\tau$, thus defining the complete form of \cref{recon}. The benefits are two fold. First, a progressive learning approach stabilizes the early portion of training, as the objective function may be less likely to become stuck at a local minimum while trying to encode and reconstruct long sequences all at once. Second, updating the parameters corresponding to only one time instance at a time has a regularizing effect by providing a more granular gradient that is less likely to interfere with the adversarial training for both the encoder and generator. The combination of both of these effects are demonstrated in \cref{fatjust}, as the training dynamics of the model are compared between using the reconstruction loss of \cref{sum_recon} and \cref{recon}. We show that the application of the objective in \cref{recon} actually minimizes the loss of \cref{sum_recon} more effectively than applying it directly in the sines dataset. The efficacy of the $\FAT_{t}$ operator can be expected to grow with longer, more complex sequences.

\subsection{Complete Model}
Time series data collected in the physical world such as sensor measurements will have some degree of stochasticity. We cannot replicate this stochasticity using a reconstruction objective alone. To this end, we introduce the feature space discriminator $d_x: \mathcal{X}^T \rightarrow \mathcal{Y}^T$ which maps sequences to classifications, such that $y_{1:T}=d(\xot)$ and $\hat{y}_{1:T} = d(\bar{x}_{1:T})$. This loss is applied to the reconstructions $\bar{x}_{1:T}$, such that the objective can be minimized both through the encoder and the generator. In the Least Squares GAN form, the objective function of the discriminator described by,
\begin{align}
\label{diseq}
&\mathcal{L}_{dx}= \frac{1}{2}\E_{\xot \sim p}(\sum_{t} \|1-y_t\|^2) + \frac{1}{2}\E_{\bar{x}_{1:T} \sim \hat{p}}(\sum_{t} \| \hat{y}_t \|^2) \\
&\mathcal{L}_{fx}= \frac{1}{2}\E_{\bar{x}_{1:T} \sim \hat{p}}(\sum_{t} \| 1-\hat{y}_t \|^2)
\label{genseq}
\end{align}

We now introduce the encoding discriminator $d_z: \mathcal{Z} \rightarrow \mathcal{Y_{Z}}$ such that $y_{z} = d_{z}(z)$ and $\hat{y}_{z} = d_{z}(z_x)$. Also in the LSGAN form,
\begin{align}
\label{diz}
& \mathcal{L}_{dz}= \frac{1}{2}\E_{z \sim p_{z}}(\|1-y_{z}\|^2) + \frac{1}{2}\E_{z_{x} \sim q} (\| \hat{y_{z}} \|^2) \\ & \mathcal{L}_{ez}=\frac{1}{2}\E_{z_{x} \sim q} (\|1- \hat{y_{z}} \|^2)
\label{giz}
\end{align}
describe the objective functions for the discriminator and encoder, respectively.

Putting everything together, there are three measures of divergence we will minimize with our complete model. The adversarial training between the objective functions described by \cref{diseq} and \cref{genseq} minimize \cref{div1} directly, where $D$ is the $\chi^2$-divergence in the LSGAN formulation. The adversarial training described by the objective functions \cref{diz} and \cref{giz} apply to the divergence of \cref{div2}, also minimizing $\chi^2$-divergence of the intermediate encodings. Finally, \cref{recon} corresponds to \cref{div3}, minimizing the Kullback–Leibler (KL) divergence through maximum likelihood (ML) supervised training. In total, all parameter optimization occurs through,
\begin{align}
\label{objective}
& \min_{e}\min_{g}(\lambda_{1} \mathcal{L}_{recon} + \mathcal{L}_{ez} + \revone{\lambda_{2}} \mathcal{L}_{fx}) \nonumber \\
& \min_{d_{z}}(\mathcal{L}_{dz}) \nonumber \\
& \min_{d_{x}}(\mathcal{L}_{dx}) \nonumber \\
\end{align}
thus describing the complete objective of the \textit{FETSGAN} architecture.

\subsection{Implementation}
\revone{The hyperparameters of the model are $\lambda_{1}$, $\lambda_{2}$, and $\epsilon$. Additionally, while  not described in notation, the dimensionality of $p_{z}$ and $p_\eta$ are also parameters of the model. In terms of tuning these values, $p_{z}$ should generally correspond to the estimated complexity of the input signals while $p_{\eta}$ should correspond to the complexity of the noise. Overestimating these values may simply lead to more complex encodings than are required, and potentially more unstable learning in the embedding space described by \cref{diz,giz}. $\lambda_{1}$ and $\lambda_{2}$ depend primarily on the scaling of the data. In our experiments, all data is normalized between $(-1,1)$. Finally, $\epsilon$ should be chosen based on the model's use-case regarding what amount of error is acceptable in reconstruction. In other words, this value represents the amount of error tolerable for the signal, where any error under this value can be considered stochastic noise subject to adversarial feature space learning.} All hyperparameters are fairly robust. To demonstrate this, the parameters are fixed for all experiments. The primary parameters are $\lambda_{1}=10$, $\lambda_{2} = 1$, and ${\epsilon}= 0.1$. The prior distribution $p_{z}$ and the noise distribution $p_\eta$ contain four dimensions, and both are sampled from $p_{z},p_{\eta} \sim \mathcal{U}(-1,1)$. All models are trained with the Adam optimization strategy \cite{kingma2014adam}, where the learning rate for the generator, encoder and both discriminators is $0.001$. These learning rates decay exponentially in the last $10\%$ of epochs. The full model implementation in Pytorch and instructions for experimental reproduction are provided in the link. \footnote{\url{https://github.com/jbeck9/FETSGAN}}

\section{Experiments}
\label{experiments}
\subsection{Experimental Setup \& Datasets}
For our experiments, we compare the performance of our model primarily against TimeGAN \cite{yoonTimeseriesGenerativeAdversarial2019} and RCGAN \cite{estebanRealvaluedMedicalTime2017}. As a baseline comparison, we have also included a purely autoregressive method that was trained using only teacher forcing. \revone{The methods of \cite{xuCOTGANGeneratingSequential2020,
jarrettTimeseriesGenerationContrastive2021, colettaConstrainedTimeSeriesGeneration2023} are omitted at the time of writing due to a lack of available implementations for reproduction.} Due to those limitations, we limit the scope of our conclusions to time series models with adversarial learning applied directly to the feature space, or low dimensional encodings of the feature space.

The efficacy of our model is shown along three dimensions. First, we demonstrate the qualitative similarity between the original data and our generated data in \cref{dismatch}. Then, we demonstrate the unique ability of our method to interpret the prior sampling distribution as a way of providing selective samples that are similar in style to specific samples from the original dataset in \cref{interp}. Finally, we demonstrate that under strenuous classification and prediction tasks, our method holds state of the art performance among adversarial learners that apply directly on the feature space for time series data in \cref{predict}. 

We use three primary datasets for analysis. First, we generate a one dimensional sines dataset of length $T=100$, where the amplitude, frequency, and phase for each sequence are sampled from a uniform distribution. We also use the six-dimensional historical Google stocks dataset, containing stock price information from 2004 to 2019 of various lengths $T$. This dataset is provided directly in the code repository. Finally, we use 6 dimensions of the UCI Appliances energy dataset \cite{candanedoDataDrivenPrediction2017} and real-valued traffic and weather data from the UCI Metro Interstate dataset \cite{houge}.

\begin{figure*}[h]
\includegraphics[width=\textwidth]{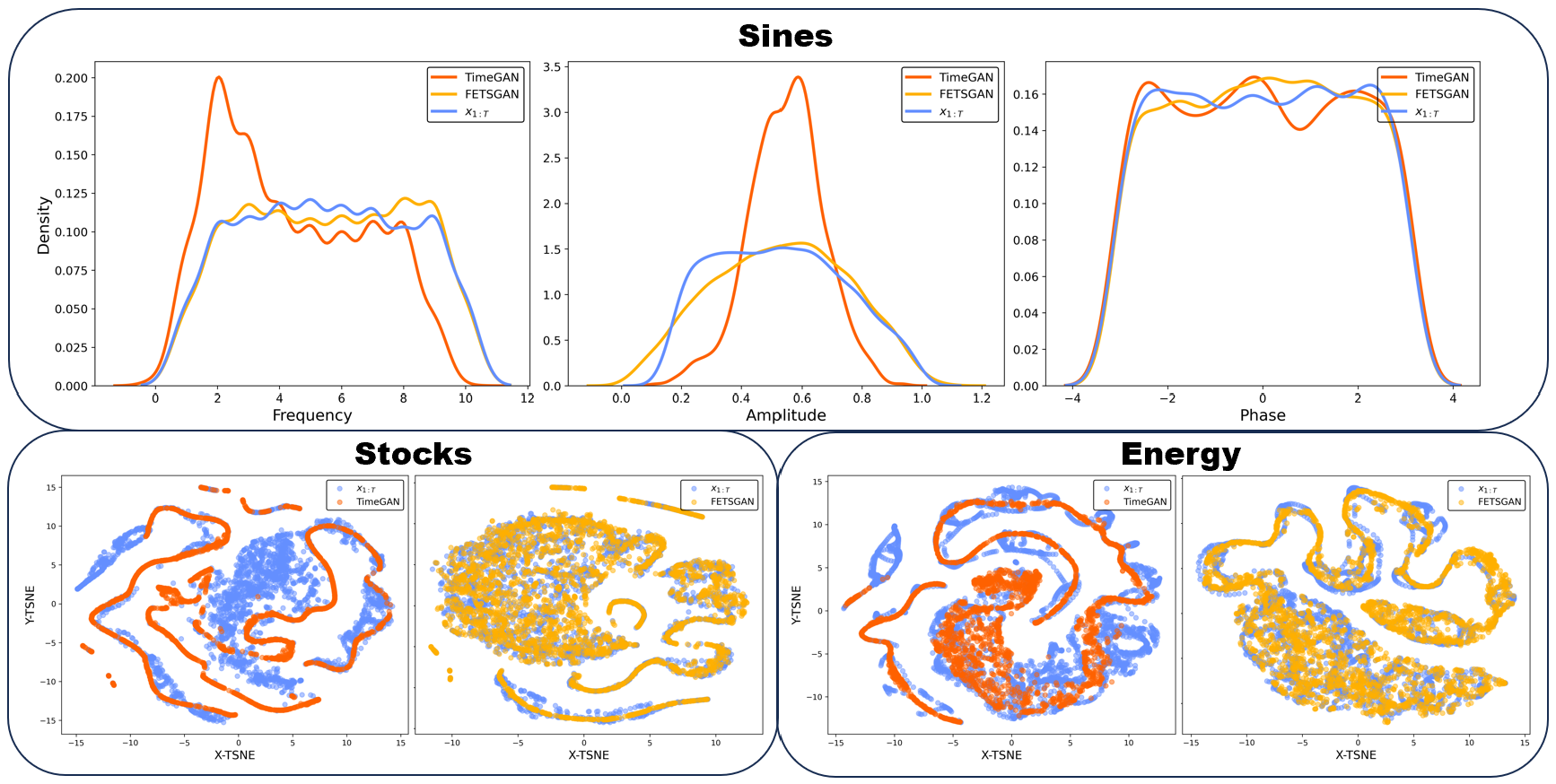}
\caption{The qualitative visualization results are shown. Sines data is shown on the top row, with the dominant component of the DFT for each sequence's frequency, amplitude, and phase (left to right) is shown as both a histogram and kernel density estimate (KDE). On the bottom row, we show a TSNE visualization for the stock dataset (left) and energy data (right), following the same procedure from \cite{yoonTimeseriesGenerativeAdversarial2019}. In all graphs, real data $x_{1:T}$ is shown in blue, synthetic data $\hat{x}_{1:T}$ (FETSGAN) is in orange, and the synthetic data generated by TimeGAN is shown in red. \textit{FETSGAN} produces data that more closely matches the original data by a substantial margin.
}
\label{datasimfig}
\centering
\end{figure*}

\subsection{Distribution Matching}
\label{dismatch}

Visualizing synthetic data generated from an adversarial learning process is important to analyze the extent of mode collapse that may have occurred. This task is tricky for time series data, as it is possible that temporal mode collapse could exist, but be obscured if the temporal dimension is flattened for analysis. In the case of the sines dataset, we can reduce the sequences to a single dimension by simply capturing the \textit{dominant frequency} in the sequence, using the Discrete Fourier Transform (DFT). Here, the dominant frequency is given by $\argmax_{f} DFT(x_{1:T})$. Since the original data consists of a sine wave with a single frequency, this provides a valid analysis of the entire sequence. The amplitude and phase of the corresponding dominant frequency are taken as well. A histogram can then be produced, comparing the distribution of frequencies, amplitudes, and phases in each dataset. For the stocks and energy datasets, we settled for a visual comparison in the flattened temporal dimension using TSNE visualization \cite{maatenVisualizingDataUsing2008}, repeating the procedure reported in \cite{yoonTimeseriesGenerativeAdversarial2019}. The results of these visualizations are shown in \cref{datasimfig}. While the results for TimeGAN generally match what is shown in \cite{yoonTimeseriesGenerativeAdversarial2019}, \textit{FETSGAN} demonstrates a substantial improvement over TimeGAN, RCGAN, and the baseline autoregressive models in matching the underlying distribution for all datasets used.

\subsection{Selective Sampling}
\label{interp}

\begin{figure*}[h]
\begin{center}
\includegraphics[width= 4in]{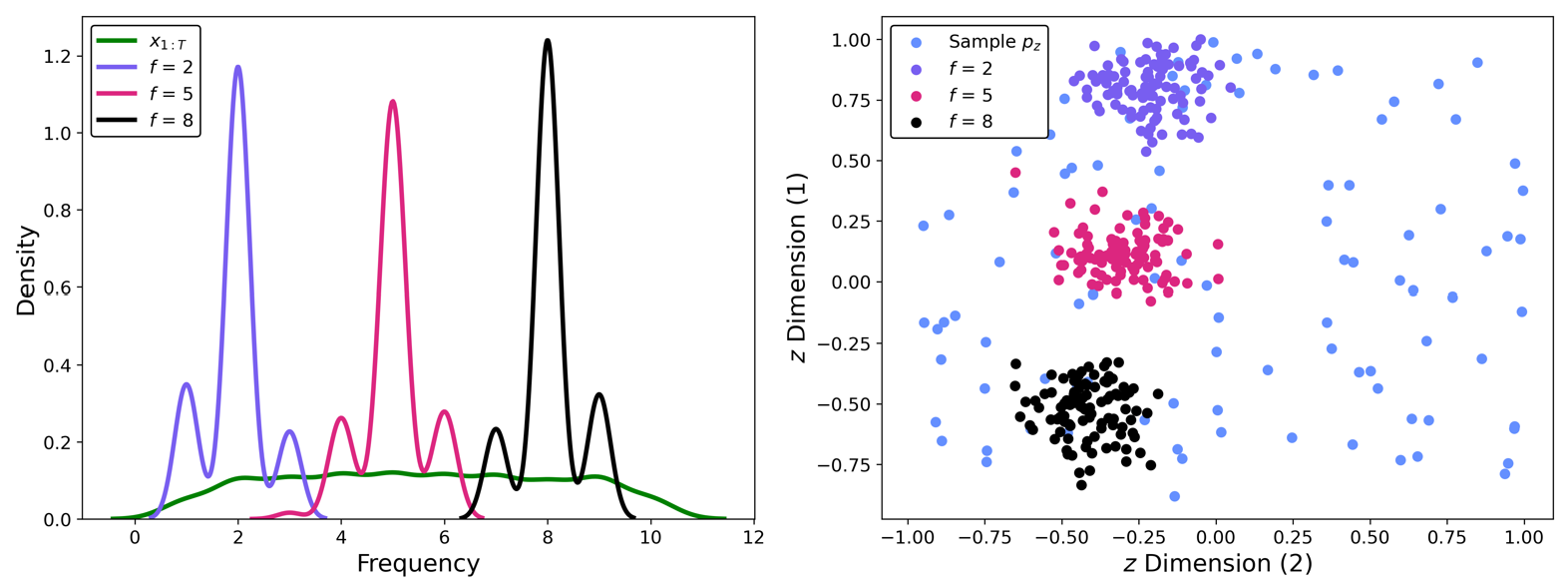}
\caption{ Selective sampling of the prior distribution $p_z$ is shown. Sines data is shown, where $100$ random samples $z_s$ were taken near the encodings of three sine waves with frequencies $f=2,5,8$ and then fed to the generator. Similar to \ref{datasimfig}, the corresponding histogram and KDE plot for  $DFT(g(z_s))$ is shown on the left. Projections of the intermediate encodings are shown on the right, demonstrating spatial interpretability.
}
\label{datasamfig}
\end{center}
\end{figure*}

There is an obvious use case that at inference time, perhaps there is a need to produce a specific \textit{style} of data, or to sample from a specific portion of the data distribution $p(X_{1:T})$. Because our model forces dimensionality reduction through an encoder that is forced to match $q$ to $p_{z}$, we can leverage spatial relationships in the latent space $z$ to selectively sample from the prior distribution $p_z$ at inference time. This allows us to produce synthetic data that retains a specific style. To demonstrate this, three sine waves were taken from the data, corresponding to sequences $x= sin(2\pi ft)$ with $f= 2,5,8$. These sequences were then encoded as $z_{x}=e(x)$. Finally, new sequences were generated by adding noise to these encodings, such that $x_{\eta}= g(z_{x} + \eta)$, where $\eta \sim \mathcal{N}(0,0.1)$. The results in \cref{datasamfig} show that, as expected, spatial relationships are maintained in the latent space $z$. We are able to produce synthetic sine waves that maintain the close relationship to the anchor point $x$ they were sampled near, allowing the ability to produce synthetic sequences within an expected range of style.

\begin{table*}[ht]
\centering
\caption{Predictive \& Discriminative Scores. Best scores shown in \textbf{bold}.}
\vskip 0.15in
\begin{tabular}[width=\textwidth]{clllll}
\hline
Model                                               & \multicolumn{1}{c}{Metric}                & \multicolumn{1}{c}{Sines} & \multicolumn{1}{c}{Energy} & \multicolumn{1}{c}{Stocks} & \multicolumn{1}{c}{Metro} \\  \hline
\multicolumn{1}{c|}{\multirow{4}{*}{\shortstack{T-Forcing \\ \tiny{(64,134 parameters)}}}}         & \multicolumn{1}{l|}{+1 Step Predict}      &$.011 \pm .004$                           &$.018 \pm .002$                            &$.047 \pm .003$                            &$.093 \pm .002$                            \\
\multicolumn{1}{c|}{}                               & \multicolumn{1}{l|}{+3 Step Predict}      &$.083 \pm .010$                           &$.028 \pm .002$                            &$.064 \pm .006$                            &$.168 \pm .004$                            \\
\multicolumn{1}{c|}{}                               & \multicolumn{1}{l|}{+5 Step Predict}     &$.421 \pm .220$                           &$.034 \pm .003$                            &$.084 \pm .014$                            &$.223 \pm .013$                            \\
\multicolumn{1}{c|}{}                               & \multicolumn{1}{l|}{Dis. Score} &$.494 \pm .008$                           &$.246 \pm .053$                            &$.243 \pm .031$                            &$.264 \pm .021$                            \\ \hline
\multicolumn{1}{c|}{\multirow{4}{*}{\shortstack{RCGAN \\ \tiny{(93,717 parameters)}}}}          & \multicolumn{1}{l|}{+1 Step Predict}      &$.011 \pm .002$                           &$.035 \pm .007$                            &$.026 \pm .001$                            &$.065 \pm .002$                            \\
\multicolumn{1}{c|}{}                               & \multicolumn{1}{l|}{+3 Step Predict}      &$.038 \pm .010$                           &$.045 \pm .007$                            &$.073 \pm .020$                            &$.134 \pm .003$                            \\
\multicolumn{1}{c|}{}                               & \multicolumn{1}{l|}{+5 Step Predict}     &$.073 \pm .020$                           &$.054 \pm .008$                            &$.094 \pm .026$                            &$.180 \pm .003$                            \\
\multicolumn{1}{c|}{}                               & \multicolumn{1}{l|}{Dis. Score} &$.467 \pm .040$                           &$.490 \pm .011$                            &$.459 \pm .018$                            &$.114 \pm .031$                            \\ \hline
\multicolumn{1}{c|}{\multirow{4}{*}{\shortstack{TimeGAN \\ \tiny{(337,858 parameters)}}}}        & \multicolumn{1}{l|}{+1 Step Predict}      &$.046 \pm .043$                           &$.021 \pm .002$                            &$.039 \pm .001$                            &$.146 \pm .038$                            \\
\multicolumn{1}{c|}{}                               & \multicolumn{1}{l|}{+3 Step Predict}      &$.133 \pm .121$                           &$.026 \pm .001$                            &$.048 \pm .002$                            &$.271 \pm .047$                            \\
\multicolumn{1}{c|}{}                               & \multicolumn{1}{l|}{+5 Step Predict}     &$.167 \pm .132$                           &$.031 \pm .001$                            &$.053 \pm .002$                            &$.330 \pm .054$                            \\
\multicolumn{1}{c|}{}                               & \multicolumn{1}{l|}{Dis. Score} &$.317 \pm .105$                           &$.164 \pm .066$                            &$.268 \pm .054$                            &$.374 \pm .104$                            \\ \hline
\multicolumn{1}{c|}{\multirow{4}{*}{\shortstack{FETSGAN-FAT \\ \tiny{(262,686 parameters)}}}}  & \multicolumn{1}{l|}{+1 Step Predict}      &$.016 \pm .004$                           &$.025 \pm .003$                            &$.030 \pm .001$                            &$.063 \pm .001$                            \\
\multicolumn{1}{c|}{}                               & \multicolumn{1}{l|}{+3 Step Predict}      &$.041 \pm .006$                           &$.029 \pm .001$                            &$.039 \pm .002$                            &$\textbf{.124} \pm .002$                            \\
\multicolumn{1}{c|}{}                               & \multicolumn{1}{l|}{+5 Step Predict}     &$.058 \pm .008$                           &$.032 \pm .001$                            &$.042 \pm .001$                            &$\textbf{.165} \pm .002$                            \\
\multicolumn{1}{c|}{}                               & \multicolumn{1}{l|}{Dis. Score} &$.249 \pm .125$                           &$.237 \pm .114$                            &$.055 \pm .032$                            &$.056 \pm .031$                            \\ \hline
\multicolumn{1}{c|}{\multirow{4}{*}{\shortstack{FETSGAN-FD \\ \tiny{(182,827 parameters)}}}}  & \multicolumn{1}{l|}{+1 Step Predict}      &$.008 \pm .005$                           &$.018 \pm .002$                            &$.031 \pm .002$                            &$.077 \pm .003$                            \\
\multicolumn{1}{c|}{}                               & \multicolumn{1}{l|}{+3 Step Predict}      &$.019 \pm .004$                           &$.026 \pm .001$                            &$.037 \pm .001$                            &$.155 \pm .008$                            \\
\multicolumn{1}{c|}{}                               & \multicolumn{1}{l|}{+5 Step Predict}     &$.028 \pm .003$                           &$.031 \pm .002$                            &$.041 \pm .002$                            &$.192 \pm .006$                            \\
\multicolumn{1}{c|}{}                               & \multicolumn{1}{l|}{Dis. Score} &$.002 \pm .007$                           &$.189 \pm .041$                            &$.086 \pm .031$                            &$.122 \pm .066$                            \\ \hline
\multicolumn{1}{c|}{\multirow{4}{*}{\shortstack{FETSGAN \\ \tiny{(262,686 parameters)}}}}       & \multicolumn{1}{l|}{+1 Step Predict}      &$\textbf{.007} \pm .006$                           &$\textbf{.016} \pm .001$                            &$\textbf{.026} \pm .001$                            &$\textbf{.062} \pm .001$                            \\
\multicolumn{1}{c|}{}                               & \multicolumn{1}{l|}{+3 Step Predict}      &$\textbf{.017} \pm .005$                           &$\textbf{.023} \pm .001$                            &$\textbf{.036} \pm .001$                            &$.128 \pm .002$                            \\
\multicolumn{1}{c|}{}                               & \multicolumn{1}{l|}{+5 Step Predict}     &$\textbf{.026} \pm .003$                           &$\textbf{.028} \pm .001$                            &$\textbf{.040} \pm .001$                            &$.169 \pm .001$                            \\
\multicolumn{1}{c|}{}                               & \multicolumn{1}{l|}{Dis. Score} &$\textbf{.001} \pm .004$                           &$\textbf{.030} \pm .012$                            &$\textbf{.005} \pm .007$                            &$\textbf{.025} \pm .014$                            \\ \hline
\end{tabular}
\label{scores}
\end{table*}
\subsection{Performance Metrics}
\label{predict}

To compare quantitative performance between models, we apply two testing metrics, \textit{discriminative score}, and \textit{predictive score}. Discriminative score is measured by training an ad hoc RNN classifier to discriminate between real dataset and a static synthetic dataset generated by each model. The best model will have the lowest score $0.5 - pred$, corresponding to how far the predictions were below the decision boundary, where a score of $0$ corresponds to indistinguishable data. In the case of prediction, the ``Train on Synthetic, Test on Real'' (TSTR) approach is used \cite{estebanRealvaluedMedicalTime2017}. A simple RNN is trained as a forecasting model, predicting 1, 3, and 5 steps into the future, given the sequence $x_{1:t}$ for any valid step where $t + step \le T$ on synthetic data. Then, MAE prediction error from the trained model is measured on the real dataset. The best model is the one which produces the lowest prediction error on real data. Whenever the model under test calls for an RNN style network, a Gated Recurrent Unit (GRU) of 64 cells and 3 layers was used. For each architecture, 3 models were trained, and sampled 5 times. Thus, 15 tests were conducted for each value in \cref{scores}. Training for all models occurred under 1000 epochs using the Adam optimizer with a learning rate of $0.001$, including the classification and prediction models. Variations of \textit{FETSGAN} are included to analyze sources of gain. \textit{FETSGAN}-FAT removes the $\FAT_{t}$ operation by replacing \cref{recon} with \cref{sum_recon}. \textit{FETSGAN}-FD trains without the feature space discriminator $d_x$. The complete model scores best in totality, and the variations of the \textit{FETSGAN} show statistically significant improvement over RCGAN and TimeGAN in all cases. We note that in the case of the noisy and lengthy Energy dataset, only the complete \textit{FETSGAN} model was able to produce data realistic enough to completely fool the ad-hoc discriminator with a score under $0.1$ for all experiments. The number of trainable parameters for each model is also shown in \cref{scores}, which was fixed for all experiments.

\revone{
\subsection{Limitations, Future Work, \& Ethical Considerations}
Fully embedding time-series data using an adversarial autoencoder ensures the synthetic data produced more closely matches the full distribution of the target data, and this auto-encoding reconstruction has a regularizing effect on the adversarial training. However, this also limits the potential applications of this methodology. Data such as video or language tokens are too highly dimensional to be reliably reduced to vector embeddings. As such, the realistic use cases are limited to real-valued time-series signals with relatively low dimensionality. The architecture chosen in this work was motivated in part by the fact that it is not auto-regressive in nature, and thus does not suffer from exposure bias. Limiting exposure bias would allow for more robust use of auto-regressive methods such as attention-based transformers, and constitutes a promising pathway for future work.

While the applications of our work may not reach the scope and magnitude of other synthetic data generation models such as large language models (LLMs), they are similar with respect to the ethical considerations highlighted in \cite{rayChatGPTComprehensiveReview2023}. Our methodology can straightforwardly be used to produce synthetic datasets where privacy is a concern within the original data. The primary responsibility for using this methodology and those like it is to ensure it is made explicitly clear that the data produced is synthetic. Additionally, the creator of the model should thoroughly ensure it sufficiently matches the distributions of the target data before applying it to problems with real-world impact.
}

\section{Conclusion}
\label{conclusion}
In this paper we introduce \textit{FETSGAN}, a novel approach to real-valued time series generation that combines feature space adversarial learning with the adversarial autoencoder framework. In addition, we introduce the $\FAT$ operator, which provides a regularizing effect on training complex temporal sequences that are produced from an intermediate encoding. Finally, the method shown here provides an interpretable latent space, allowing higher flexibility for selective sampling at inference time. We demonstrate significant improvement over current adversarial methods applied directly to the feature space or encodings thereof. In future work, we intend to leverage the accuracy and interpretability of this model on a variety of datasets to demonstrate real world utility for synthetic data to aid in various applied machine learning models in forecasting and classification.

\section{Competing Interests, Author Contribution, \& Data Availability}
\label{ci}
\subsection{Competing Interests}
The authors declare no known competing interests in the content of this manuscript. Funding for this work was partially provided by the Collaborative Sciences Center for Road Safety (CSCRS), as well as the University of Tennessee, Knoxville.


\subsection{Data Availability and Ethical Use}
All the data used for experimentation was publicly available, and contains no sensitive or personal information of any kind. No original datasets were produced through this research. All datasets are either provided through citation, or provided directly at the linked repository. 


\bibliography{paper, NCAA_rev}

\begin{thebibliography}{10}
\providecommand{\url}[1]{#1}
\csname url@samestyle\endcsname
\providecommand{\newblock}{\relax}
\providecommand{\bibinfo}[2]{#2}
\providecommand{\BIBentrySTDinterwordspacing}{\spaceskip=0pt\relax}
\providecommand{\BIBentryALTinterwordstretchfactor}{4}
\providecommand{\BIBentryALTinterwordspacing}{\spaceskip=\fontdimen2\font plus
\BIBentryALTinterwordstretchfactor\fontdimen3\font minus
  \fontdimen4\font\relax}
\providecommand{\BIBforeignlanguage}[2]{{%
\expandafter\ifx\csname l@#1\endcsname\relax
\typeout{** WARNING: IEEEtran.bst: No hyphenation pattern has been}%
\typeout{** loaded for the language `#1'. Using the pattern for}%
\typeout{** the default language instead.}%
\else
\language=\csname l@#1\endcsname
\fi
#2}}
\providecommand{\BIBdecl}{\relax}
\BIBdecl

\bibitem{estebanRealvaluedMedicalTime2017}
C.~Esteban, S.~L. Hyland, and G.~R{\"a}tsch, ``Real-valued ({{Medical}}) {{Time
  Series Generation}} with {{Recurrent Conditional GANs}},'' Dec. 2017.

\bibitem{mogrenCRNNGANContinuousRecurrent2016}
O.~Mogren, ``C-{{RNN-GAN}}: {{Continuous}} recurrent neural networks with
  adversarial training,'' Nov. 2016.

\bibitem{makhzaniAdversarialAutoencoders2016}
A.~Makhzani, J.~Shlens, N.~Jaitly, I.~Goodfellow, and B.~Frey, ``Adversarial
  {{Autoencoders}},'' May 2016.

\bibitem{sutskeverSequenceSequenceLearning2014a}
I.~Sutskever, O.~Vinyals, and Q.~V. Le, ``Sequence to {{Sequence Learning}}
  with {{Neural Networks}},'' Dec. 2014.

\bibitem{yoonTimeseriesGenerativeAdversarial2019}
J.~Yoon, D.~Jarrett, and M.~{van der Schaar}, ``Time-series {{Generative
  Adversarial Networks}},'' in \emph{Advances in {{Neural Information
  Processing Systems}}}, vol.~32.\hskip 1em plus 0.5em minus 0.4em\relax
  {Curran Associates, Inc.}, 2019.

\bibitem{bengioScheduledSamplingSequence2015}
S.~Bengio, O.~Vinyals, N.~Jaitly, and N.~Shazeer, ``Scheduled {{Sampling}} for
  {{Sequence Prediction}} with {{Recurrent Neural Networks}},'' Sep. 2015.

\bibitem{lambProfessorForcingNew2016}
A.~Lamb, A.~Goyal, Y.~Zhang, S.~Zhang, A.~Courville, and Y.~Bengio, ``Professor
  {{Forcing}}: {{A New Algorithm}} for {{Training Recurrent Networks}},'' Oct.
  2016.

\bibitem{chungEmpiricalEvaluationGated2014}
J.~Chung, C.~Gulcehre, K.~Cho, and Y.~Bengio, ``Empirical {{Evaluation}} of
  {{Gated Recurrent Neural Networks}} on {{Sequence Modeling}},'' Dec. 2014.

\bibitem{arjovskyWassersteinGAN2017}
M.~Arjovsky, S.~Chintala, and L.~Bottou, ``Wasserstein {{GAN}},'' Dec. 2017.

\bibitem{fekriGeneratingEnergyData2020}
M.~N. Fekri, A.~M. Ghosh, and K.~Grolinger, ``Generating {{Energy Data}} for
  {{Machine Learning}} with {{Recurrent Generative Adversarial Networks}},''
  \emph{Energies}, vol.~13, no.~1, p. 130, Jan. 2020.

\bibitem{xuCOTGANGeneratingSequential2020}
T.~Xu, L.~K. Wenliang, M.~Munn, and B.~Acciaio, ``{{COT-GAN}}: {{Generating
  Sequential Data}} via {{Causal Optimal Transport}},'' in \emph{Advances in
  {{Neural Information Processing Systems}}}, vol.~33.\hskip 1em plus 0.5em
  minus 0.4em\relax {Curran Associates, Inc.}, 2020, pp. 8798--8809.

\bibitem{jarrettTimeseriesGenerationContrastive2021}
D.~Jarrett, I.~Bica, and M.~{van der Schaar}, ``Time-series {{Generation}} by
  {{Contrastive Imitation}},'' in \emph{Advances in {{Neural Information
  Processing Systems}}}, vol.~34.\hskip 1em plus 0.5em minus 0.4em\relax
  {Curran Associates, Inc.}, 2021, pp. 28\,968--28\,982.

\bibitem{colettaConstrainedTimeSeriesGeneration2023}
A.~Coletta, S.~Gopalakrishnan, D.~Borrajo, and S.~Vyetrenko, ``On the
  {{Constrained Time-Series Generation Problem}},'' \emph{Advances in Neural
  Information Processing Systems}, vol.~36, pp. 61\,048--61\,059, Dec. 2023.

\bibitem{dashMedicalTimeSeriesData2020}
S.~Dash, A.~Yale, I.~Guyon, and K.~P. Bennett, ``Medical {{Time-Series Data
  Generation Using Generative Adversarial Networks}},'' in \emph{Artificial
  {{Intelligence}} in {{Medicine}}}, ser. Lecture {{Notes}} in {{Computer
  Science}}, M.~Michalowski and R.~Moskovitch, Eds.\hskip 1em plus 0.5em minus
  0.4em\relax {Cham}: {Springer International Publishing}, 2020, pp. 382--391.

\bibitem{liCausalRecurrentVariational2023}
H.~Li, S.~Yu, and J.~Principe, ``Causal {{Recurrent Variational Autoencoder}}
  for {{Medical Time Series Generation}},'' \emph{Proceedings of the AAAI
  Conference on Artificial Intelligence}, vol.~37, no.~7, pp. 8562--8570, Jun.
  2023.

\bibitem{chattorajImprovingStabilityAdversarial2021}
S.~Chattoraj, S.~Pratiher, S.~Pratiher, and H.~Konik, ``Improving {{Stability}}
  of {{Adversarial Li-ion Cell Usage Data Generation}} using {{Generative
  Latent Space Modelling}},'' in \emph{{{ICASSP}} 2021 - 2021 {{IEEE
  International Conference}} on {{Acoustics}}, {{Speech}} and {{Signal
  Processing}} ({{ICASSP}})}, Jun. 2021, pp. 8047--8051.

\bibitem{fochesatoUseConditionalTimeGAN2022}
M.~Fochesato, F.~Khayatian, D.~F. Lima, and Z.~Nagy, ``On the use of
  conditional {{TimeGAN}} to enhance the robustness of a reinforcement learning
  agent in the building domain,'' in \emph{Proceedings of the 9th {{ACM
  International Conference}} on {{Systems}} for {{Energy-Efficient Buildings}},
  {{Cities}}, and {{Transportation}}}, ser. {{BuildSys}} '22.\hskip 1em plus
  0.5em minus 0.4em\relax {New York, NY, USA}: {Association for Computing
  Machinery}, Dec. 2022, pp. 208--217.

\bibitem{adibSyntheticECGSignal2023}
E.~Adib, A.~S. Fernandez, F.~Afghah, and J.~J. Prevost, ``Synthetic {{ECG
  Signal Generation Using Probabilistic Diffusion Models}},'' \emph{IEEE
  Access}, vol.~11, pp. 75\,818--75\,828, 2023.

\bibitem{vaswaniAttentionAllYou2017}
A.~Vaswani, N.~Shazeer, N.~Parmar, J.~Uszkoreit, L.~Jones, A.~N. Gomez,
  {\L}.~ukasz Kaiser, and I.~Polosukhin, ``Attention is {{All}} you {{Need}},''
  in \emph{Advances in {{Neural Information Processing Systems}}},
  vol.~30.\hskip 1em plus 0.5em minus 0.4em\relax Curran Associates, Inc.,
  2017.

\bibitem{rayChatGPTComprehensiveReview2023}
P.~P. Ray, ``{{ChatGPT}}: {{A}} comprehensive review on background,
  applications, key challenges, bias, ethics, limitations and future scope,''
  \emph{Internet of Things and Cyber-Physical Systems}, vol.~3, pp. 121--154,
  Jan. 2023.

\bibitem{lyuImprovingClinicalPredictions2018}
X.~Lyu, M.~Hueser, S.~L. Hyland, G.~Zerveas, and G.~Raetsch, ``Improving
  {{Clinical Predictions}} through {{Unsupervised Time Series Representation
  Learning}},'' Dec. 2018.

\bibitem{daiSemisupervisedSequenceLearning2015}
A.~M. Dai and Q.~V. Le, ``Semi-supervised {{Sequence Learning}},'' Nov. 2015.

\bibitem{bianchiLearningRepresentationsMultivariate2019}
F.~M. Bianchi, L.~Livi, K.~{\O}. Mikalsen, M.~Kampffmeyer, and R.~Jenssen,
  ``Learning representations for multivariate time series with missing data
  using {{Temporal Kernelized Autoencoders}},'' Jul. 2019.

\bibitem{kingmaAutoEncodingVariationalBayes2022}
D.~P. Kingma and M.~Welling, ``Auto-{{Encoding Variational Bayes}},'' Dec.
  2013.

\bibitem{tolstikhinWassersteinAutoEncoders2019}
I.~Tolstikhin, O.~Bousquet, S.~Gelly, and B.~Schoelkopf, ``Wasserstein
  {{Auto-Encoders}},'' Dec. 2019.

\bibitem{liuUnsupervisedImagetoImageTranslation2017}
M.-Y. Liu, T.~Breuel, and J.~Kautz, ``Unsupervised {{Image-to-Image Translation
  Networks}},'' in \emph{Advances in {{Neural Information Processing
  Systems}}}, vol.~30.\hskip 1em plus 0.5em minus 0.4em\relax {Curran
  Associates, Inc.}, 2017.

\bibitem{zhu2017unpaired}
J.-Y. Zhu, T.~Park, P.~Isola, and A.~A. Efros, ``Unpaired image-to-image
  translation using cycle-consistent adversarial networks,'' in
  \emph{Proceedings of the IEEE international conference on computer vision},
  2017, pp. 2223--2232.

\bibitem{maoLeastSquaresGenerative2017}
X.~Mao, Q.~Li, H.~Xie, R.~Y.~K. Lau, Z.~Wang, and S.~P. Smolley, ``Least
  {{Squares Generative Adversarial Networks}},'' Apr. 2017.

\bibitem{miyatoSpectralNormalizationGenerative2018}
T.~Miyato, T.~Kataoka, M.~Koyama, and Y.~Yoshida, ``Spectral {{Normalization}}
  for {{Generative Adversarial Networks}},'' Feb. 2018.

\bibitem{xiangEffectsBatchWeight2017}
S.~Xiang and H.~Li, ``On the {{Effects}} of {{Batch}} and {{Weight
  Normalization}} in {{Generative Adversarial Networks}},'' Dec. 2017.

\bibitem{kingma2014adam}
D.~P. Kingma and J.~Ba, ``Adam: A method for stochastic optimization,''
  \emph{arXiv preprint arXiv:1412.6980}, 2014.

\bibitem{candanedoDataDrivenPrediction2017}
L.~M. Candanedo, V.~Feldheim, and D.~Deramaix, ``Data driven prediction models
  of energy use of appliances in a low-energy house,'' \emph{Energy and
  Buildings}, vol. 140, pp. 81--97, Apr. 2017.

\bibitem{houge}
J.~Hogue, ``Traffic data from mn department of transportation, weather data
  from openweathermap,'' 2018.

\bibitem{maatenVisualizingDataUsing2008}
L.~van~der Maaten and G.~Hinton, ``Visualizing {{Data}} using t-{{SNE}},''
  \emph{Journal of Machine Learning Research}, vol.~9, no.~86, pp. 2579--2605,
  2008.

\end{thebibliography}
\bibliographystyle{ieee}



\end{document}